\documentclass[letterpaper, 10 pt, conference]{ieeeconf}

\IEEEoverridecommandlockouts

\usepackage{color}
\usepackage{graphicx}
\usepackage{algpseudocode}
\usepackage{algorithm}
\usepackage{amsmath, amssymb}
\usepackage{framed}
\usepackage{tikz}
\usepackage{pgfplots}

\usepackage{graphicx}
\usepackage{subfig}

\newcommand{\argmin}[1]{\underset{#1}{\operatorname{arg}\operatorname{min}}\;}

\renewcommand{\v}{\boldsymbol}
\renewcommand{\t}[1]{{\textrm{#1}}}
\newcommand{\mat}{\boldsymbol}
\newcommand{\R}{\mathbb{R}}

\newcommand{\data}{\mathcal{D}}
\newcommand{\eq}{Equation~} 

\newcommand{\fig}{Figure~}
\newcommand{\alg}{Algorithm~}

\newcommand{\sect}{Section~}

\newcommand{\iIdx}{^{(i)}}

\newcommand{\idx}[1]{^{(#1)}}
\newcommand{\opt}{^\star}
\newcommand{\trans}{^\top}
\newcommand{\inv}{^{-1}}

\newcommand{\period}{~.\quad}
\newcommand{\pc}{\mat P}
\newcommand{\depth}{\mat D}

\newcommand{\nn}{f}
\newcommand{\model}{\v m}

\ifdefined\success
\renewcommand{\success}{S}
\else
\newcommand{\success}{S}
\fi

\newcommand{\traj}{\bar{\v x}}
\newcommand{\trajI}{\traj_{0:T}}

\newcommand{\parA}{\v\theta}
\newcommand{\parB}{\v w}

\newcommand{\parF}{\v \gamma}

\newcommand{\parH}{\v \beta}

\newcommand{\etal}{\textit{et al}. }

\newcommand{\specialcell}[2][c]{
	\begin{tabular}[#1]{@{}c@{}}#2\end{tabular}}

\newcommand{\specialcellL}[2][c]{
	\begin{tabular}[#1]{@{}l@{}}#2\end{tabular}}

\ifdefined\newcolumntype
\newcolumntype{?}{!{\vrule width 1.3pt}}
\fi

\newlength\figureheight
\newlength\figurewidth

\title{\LARGE \bf
Kinematic Morphing Networks for Manipulation Skill Transfer
}

\author{Peter Englert and Marc Toussaint
\thanks{\hspace{-3mm}\rule{.48\textwidth}{0.4pt} \mbox{\hfill} \hspace{-3mm}\mbox{This work was supported by the DFG project Exploration Challenge} \mbox{TO 409/9-1.}\vspace{1mm}}
\thanks{\hspace{-3mm}\mbox{Peter Englert and Marc Toussaint are with the Machine Learning \&} \mbox{Robotics Lab, University of Stuttgart, Germany.}        \mbox{Email: englertpr@gmail.com}}
}

\begin{document}

\maketitle
\thispagestyle{empty}
\pagestyle{empty}

\begin{abstract} 
The transfer of a robot skill between different geometric environments is non-trivial since a wide variety of environments exists, sensor observations as well as robot motions are high-dimensional, and the environment might only be partially observed.
We consider the problem of extracting a low-dimensional description of the manipulated environment in form of a kinematic model.
This allows us to transfer a skill by defining a policy on a prototype model and morphing the observed environment to this prototype.
A deep neural network is used to map depth image observations of the environment to morphing parameter, which include transformation and configuration parameters of the prototype model.
Using the concatenation property of affine transformations and the ability to convert point clouds to depth images allows to apply the network in an iterative
manner.
The network is trained on data generated in a simulator and on augmented data that is created by using network predictions.
The algorithm is evaluated on different tasks, where it is shown that iterative predictions lead to a higher accuracy than one-step predictions.
\end{abstract}

\section{Introduction} 
\label{sec:introduction}  
Modern robots are equipped with sensors (e.g., camera, laser scanner) that allow them to perceive their environment.
For many policy representations, raw sensor signals are not directly usable as an input since they are too abstract and high-dimensional.
Therefore, algorithms are necessary to extract a representation that is a suitable input for such policies.
We propose to use kinematic models of the environment as such a representation.
The main objective of this work is to extract these parameters from an observed environment.

In this paper, we consider the scenario where a robot observes a manipulation environment with a depth sensor.
The observation is taken from a specific viewpoint, which often results in a measurement that only covers certain parts of the environment.
The goal is to learn a deep neural network that maps observations to parameters, which are the input to a manipulation policy.
We assume to have the kinematic model structure of the manipulated environment and a simulator that can create large amounts of supervised training data.
In the following, we describe the different components of our approach.
\begin{figure}[t]
	\centering
	\includegraphics[width=.36\textwidth]{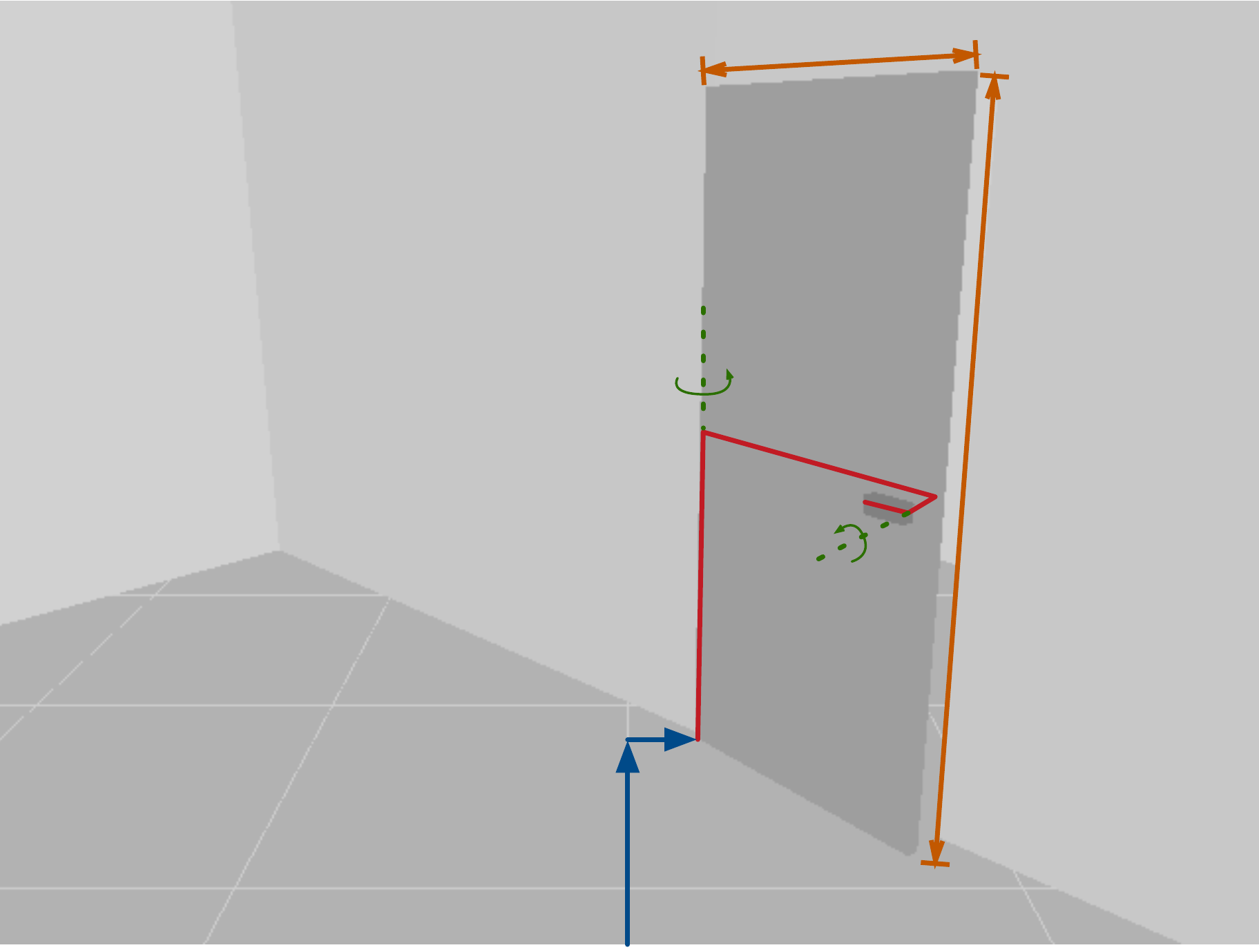}
	\caption{Kinematic model of a door that we want to extract from depth images and use to transfer manipulation skills.}
	\label{fig:door_setup}
	\vspace{-1.5em}
\end{figure}

\subsection{Environment Parametrization} 
\label{sub:parametrization} 
We represent the manipulated environment with a kinematic model that consists of rigid bodies and joints.
\fig \ref{fig:door_setup} shows such a kinematic model for a door environment with its parametrization (e.g., width, position, handle location).
A key element of our approach is a \emph{prototype} that serves as a reference to describe a model.
The models are parametrized by \emph{morphing parameters} that define the mapping of each point of a model to a corresponding point on the prototype.
The parameters of this morphing include 3D transformation parameters as well as configuration parameters of the prototype, such as the height of a door handle.
We assume that if a policy for the prototype model and the morphing parameters of an observed environment are known, then the policy can be transferred from the prototype to the observed environment.
Specifically, we will use a trajectory optimization method to define costs and constraints depending on the morphing parameter.
These costs and constraints describe how the robot should interact with the environment (e.g., contacts) in order manipulate it into a
desired state.

\subsection{Kinematic Morphing Model} 
\label{sub:model} 
The goal of this work is to extract the morphing parameters from sensor observations.
The proposed \emph{kinematic morphing model} is defined as a convolutional neural network that maps depth images to morphing parameters.
We propose a data augmentation method that uses the network predictions to generate more data.
This augmentation is done by applying the predicted morphing parameters on the input point cloud and generating a new depth image from it.
We use the same mechanism to make predictions with the network in an iterative manner.
This can be viewed as a controller that changes the inputs in multiple steps until a steady point is reached.
In our case the steady point is the prototype model and the goal is to transform all observed models to this prototype.
The advantage is that the model does not have to predict the morphing parameter in a single step. We show in our experiments that this results in an increase of the prediction accuracy.

\subsection{Data Generation in Simulation} 
\label{sub:data_gen} 
Training the parameters of a deep neural networks requires a large amount of data, which is difficult to collect in the real world since the labels would have to be provided by hand.
In this work, we follow a recent trend to generate synthetic data with simulators (see \sect \ref{sub:simulated_environments}).
A kinematic engine and OpenGL renderer are used to create 3D representations of environments for a given set of morphing parameters.
This allows to generate a large supervised dataset that consists of point clouds, depth images and morphing parameters.
In this paper, we focus on the morphing of the kinematic models and therefore assume that the background has already been removed in the data.
The kinematic morphing model is trained on this data and on the augmented data created with the kinematic morphing model.
\\\\
Combining all these ingredients provides us with a tool to extract a compact representation from high dimensional sensor data, which can be used to transfer
robot skills between different environments.
We will demonstrate these transfer abilities in the experimental section, where the same skill policy is used to manipulate doors of different shapes and
locations.
The main contribution of this work is the kinematic morphing network that is trained iteratively by augmenting the data with its own predictions.

\section{Related Work} 
\label{sec:related_work} 
\subsection{Extracting 3D Models from Data} 
\label{sub:extracting_3d_models} 
Point set registration methods try to find the transformation between two observations of a model \cite{chen1992object, gold1998}.
Iterative Closest Point (ICP) is a widely used algorithm to align two point clouds \cite{chen1992object}. ICP iterates between the steps: 1) Matching the points between the two point clouds by finding the closest pairs; 2) Computing a transformation that minimizes the distances between the point pairs; 3) Applying the transformation on a point cloud and continuing with step 1.
The advantage of these kinds of methods is that they do not require an expensive training procedures like neural networks. 
However, they still have open parameters that influence the performance and the initial estimate of the transformation is important.
The main difference to our approach is that we can also handle kinematic model parameters beyond affine transformations.
We compare our approach to ICP in the experiment in \sect \ref{ssec:evaluation}.

An alternative strategy is to extract models from motions of the environment \cite{sturm2011probabilistic, martin2014online}.
Martin-Martin \etal \cite{martin2014online} do a feature based approach by tracking the motion of different feature points and extracting a kinematic model from them.
Sturm \etal \cite{sturm2011probabilistic} follow a probabilistic approach to extract a kinematic model from pose trajectories of rigid bodies.
The objective of these algorithms is similar to our work.
The main difference to our work is that we do not require object motions, which is difficult to produce in an automated way when the environment is initially unknown.

Zhou \etal \cite{zhou2016} follow a similar approach to ours and learn a deep neural network that predicts the configuration of a kinematic hand model from depth images.  
The forward kinematic function is integrated as final layer into the network and outputs the location of each joint.
The loss function is defined on the location of these joints and an additional loss term ensures that the predicted parameter fulfil physical constraints.
In our approach, we directly compute the error in the parameter space and also train the network to predict the model structure (e.g., finger length).

\subsection{Learning in Simulated Environments} 
\label{sub:simulated_environments} 
Using simulation environments is a way to bypass the lack of large datasets in robotics. 
However, bridging the gap between simulated and real sensor data (e.g., images) is still an open research question.
In \cite{bousmalis2017arxiv}, a robot grasping skill is improved by using synthetic data generated with a simulator.
The proposed approach uses domain adaptation techniques \cite{bousmalis2017unsupervised, ganin2016domain} that map synthetic images to realistically looking
images.
They use generative adversarial networks \cite{goodfellow2014generative} to learn this mapping by using two networks that are trained adversarial.
The generated data is used to train a network that maps RGB images and actions to grasp success probabilities \cite{levine2016learning}.
Their results show that the use of simulation data leads to a consistent improvement of the overall grasp success rate.
Rusu \etal \cite{rusu2017} transfer policies from simulated to real robots by using Progressive Neural Networks \cite{Rusu2016a}.
The first column of the progressive neural network is trained in simulation. All further layers are trained on the real robot while the first column is kept fixed.
The training is done with the Asynchronous Advantage Actor-Critic \cite{mnih_2016} method. 
The input to the network is an RGB image and the outputs are a discrete velocity signal for each joint plus a value function.
Instead of following an end-to-end approach, we propose a more structured way of transferring skills by extracting a representation that is suitable as input
for standard planning methods.

Mitash \etal \cite{Mitash2017} propose an object detection algorithm based on a physics simulation and a real-world self learning mechanism.
The physics simulator uses CAD models of the objects to generate realistic scenes.
Each scene is rendered from multiple perspectives and the produced RGB-D images are used to train a convolutional neural network.
The trained network is used to generate more labeled training data in a real world environment.
Thereby, a robot is used to arrange the scene and take multiple images from different perspectives.
The detected objects with high confidence are used to create corresponding labels for all perspectives.
Their results show that the simulation part of the algorithm provide the policy with a good starting point for the real-world self learning.

\subsection{Integrating 3D Geometry in Neural Networks} 
\label{sub:integrating_3d_data_in_neural_networks} 
There are different ways of how neural networks can be used with 3D sensor data and how transformations or rendering operations can be represented with the network.
A problem that often occurs is that the observations are taken from a specific viewpoint, which leads to only partial observations of objects.
Eitel \etal \cite{eitel15iros} do object detection based on RGB-D data.
The model consists of a two-stream convolutional neural network with an RGB image input and a depth image input.
The output is a probability of how likely an object is in the image.
They compare different ways to represent depth images and to transfer pre-trained networks trained on RGB data.
Byravan \etal \cite{byravan2017} learn rigid body motions with deep neural networks based on depth data.
The model inputs are a point cloud shaped as an XYZ image and a force vector.
The network combines both inputs in a late fusion architecture and outputs a transformed point cloud image.
The transformation is done by predicting a fixed amount of object masks and corresponding rigid body transformations.

Spatial Transformer Networks (STN) are a network module to transforms an input feature map to an output feature map \cite{jaderberg2015spatial}.
STN can be used as a layer in a network and are differentiable.
The parameters of the affine transformation are predicted based on the input feature map.
Using STN leads to invariance regarding translation, scale and rotation.
Rezende \etal \cite{rezende2016} use STN to extract 3D structure from images.
A conditional latent variable model with a low dimensional codec is used to map observed data to an abstract code that a decoder maps to volume representations.
Similar to our approach, they also use an OpenGL renderer to convert from a 3D representation to image. In our case, the low dimensional representation are
interpretable kinematic parameters that can be used for planning methods.
Discretizing the 3D space might be suitable for some tasks like object detection.
However, the achieved accuracy strongly depends on the resolution of the grid.

\section{Model \& Policy Representation} 
\label{sec:representation} 
\subsection{Kinematic Model of the Manipulated Environment} 
\label{ssec:environment_representation} 
We introduce a parametrized kinematic model of the environment \mbox{$\model(\parA,\parF) \subset \R^3$}, which is defined as a set of points of the manipulated environment (e.g., a door) with parameters $\parA\in\R^n$ and $\parF\in\R^m$.
A prototype model $\model(\parA_0,\parF_0)$ is defined as a reference for other models, where $\parF_0$ and $\parA_0$ are usually set to $\v 0$.
In this paper, the term \emph{morphing} is used to describe a mapping of a model $\model(\parA,\parF)$ to the prototype $\model(\parA_0,\parF_0)$.
The parameters of the kinematic model are:

\begin{enumerate}
	\item The \emph{transformation parameter} $\parA$ of an affine transformation $\v T_{\parA} \in \t{Aff}(3)$ that describes the linear mapping between two
models.
The parameter $\parA$ are specific rotation, translation, and scale parameters around or along a certain axis.

\item The \emph{configuration parameter} $\parF$ describe the nonlinear mapping between two models that cannot be represented with an affine transformation of
the complete model.
An example is the relative position between two bodies of the environment.

\end{enumerate}
The affine transformation of the morphing operation for a given configuration $\parF$ is
\begin{equation}
	\label{eq:model_transform}
	 \model(\parA_0, \parF) = \v T_{\parA}\inv \model(\parA, \parF)\period
\end{equation}
This equation describes how all points of a model paramerized by $\parA$ transform onto a prototype $\parA_0$. 
The goal of this paper is to predict model parameter $(\parA, \parF)$ from sensor observations of the model.
These parameters relate the current observed model to the prototype model, which will be used to adapt the policy from the prototype to the observed model.

\subsection{Constrained Trajectory Optimization Policies} 
\label{sub:policy_representation} 
Our policy representation is a constrained trajectory optimization problem consisting of costs and constraints that describe how the robot should interact with the environment.
We use k-order Markov optimization \cite{17-toussaint-Newton} that finds trajectories $\trajI\in\R^{Q\times (T+1)}$ by solving the problem
\begin{align}
	\label{eq:trajloss}
	\trajI\opt = \argmin{\trajI} & \parB\trans\v\Phi^2 (\trajI, \model)\\
	\text{s.t.}\quad & \v g(\trajI,\model) \leq \v 0 \nonumber\\
	& \v h(\trajI,\model) = \v 0 \nonumber \period
\end{align}
$\v\Phi$ are cost features (e.g., endeffector position/orientation) of the trajectory and $\parB$ are feature weights. Inequality constraints
$\v g$ and equality constraints $\v h$ are used to define further properties (e.g., contacts, collision avoidance) of the motion.
The kinematic model $\model$ is an input to the cost and constraint functions, which generalizes the skill to different models.

\section{Kinematic Morphing Networks} 
\label{sec:methodology} 
In this section, we propose an approach to extract morphing parameters from environment observations.
The robot observes an instance of the kinematic model $\model(\parA,\parF)$ in form of a depth image $\depth\in\R^{W\times H}$ and corresponding point cloud
$\pc\in\R^{3\times(WH)}$.
We define the function 
\begin{align}
	f :~ \R^{W\times H} \rightarrow \R^n\times\R^m
	\label{eq:network}
\end{align}
that maps depth images $\depth$ to morphing parameter $(\parA, \parF)$ (see \sect \ref{ssec:environment_representation}).
In this paper, $f (\depth; \parH)$ is represented as a neural network with parameters $\parH\in\R^{B}$.
In the proposed approach, data of the form $\data=(\depth\iIdx, \pc\iIdx, \parA\iIdx, \parF\iIdx)_{i=1}^N$ is used to optimize the network parameters $\parH$.
In the following sections we describe the network prediction, training, and architecture.

\subsection{Iterative Network Predictions} 
\label{sub:iterative_network_predictions} 
We introduce an iterative network prediction mechanism that applies the network in \eq \eqref{eq:network} repeatedly.
For a given input $(\depth, \pc)$, the predictions are computed by iterating:
\begin{enumerate}
	\item Predicting parameters $\parA$ for $\depth$ with \eq \eqref{eq:network}.
	\item Applying the transformation $\v T_{\parA}$ to the point cloud $\pc$.
	\item Rendering a new depth image $\depth$ from $\pc$.
\end{enumerate}
These three steps are repeated until a fixed point is reached. 
The resulting point cloud after $t$ iterations is
\begin{equation} 
	\label{eq:pc_transform}
	\pc_t = \v T_{\parA_t}\inv \dots \v T_{\parA_2}\inv \v T_{\parA_1}\inv \pc~.
\end{equation} 
The idea is that in each step the point cloud is transformed a bit closer towards the prototype. 
After convergence, this point cloud should overlay with the prototype $\parA_0$.
The necessary steps are summarized in \alg \ref{alg:prediction}.
The inputs are a depth image $\depth$, a point cloud $\pc$, a previous transformation $\parA$ (by default $\parA_0$), network parameter $\parH$ and number of predictions $N$. 
In the first step of the loop, the network predicts a transformation parameter for the given depth image. 
Afterwards, the corresponding point cloud is transformed with the predicted transformation, which is then mapped to a new depth image.
Finally, the predicted transformations are concatenated that we express with the symbol $\circ$.
This procedure is repeated $N$ times. The output are a new depth image and point cloud with the corresponding morphing parameter.
An alternative to the fixed number of iterations $N$ would be to repeat the steps until the network predicts a transformation that is close to the identity transformation, which indicates convergence.
The algorithm requires a converting functionality that renders a depth image $\depth$ from the transformed point cloud $\pc$. 
The configuration parameter $\parF$ cannot be predicted in an iterative manner and only the last prediction of the network is used.
\begin{figure}[t]
	\begin{minipage}[t]{0.5\textwidth}
\begin{algorithm}[H]
\begin{algorithmic}
	\State \textbf{\emph{function}} predict$(\depth,\pc,\parA,\parH,N):$
	\State $\quad\t{\textbf{for} } d=1:N$
	\State $\qquad(\bar\parA, \parF) = \nn(\depth;\parH)$
	\State $\qquad\pc =  \v T_{\bar\parA}\inv \pc$
	\State $\qquad\depth = \t{pointCloudToDepth}(\pc)$
	\State $\qquad\parA = \bar\parA\inv \circ \parA$
	\State $\quad\t{\textbf{end}}$\\
	\vspace{.3\baselineskip}
	$\quad$\Return $(\depth,\pc,\parA,\parF)$
\end{algorithmic}
\caption{\!\!\textbf{:} Multi-step Network Predictions}
\label{alg:prediction}
\end{algorithm}
\end{minipage}
\vspace{-1.5em}
\end{figure}

\subsection{Data Generation and Network Training} 
\label{sub:data_generation_and_network_training} 
\alg \ref{alg:training} shows the combined data generation and network training.
Throughout the training, the current state of the network is used to augment the training data by using \alg \ref{alg:prediction}.
The inputs are the parametrized model $\model$ and limits of the model parameter.
In the first part of the algorithm, a dataset is generated with an OpenGL renderer that creates for a given model $\model(\parA,\parF)$ a depth image $\depth$
and a point cloud $\pc$.
The parameters are thereby sampled uniformly in the feasible parameter range defined by $L^{\t{up}}$ and $L^{\t{low}}$.
Afterwards, the network is trained on the generated dataset.
In the second phase, the trained model is used to augment the dataset by applying the model on a subset $N_\t{aug}$ of the initial data.
Thereby, the iterative network predictions with $N_\t{pred}$ predictions from \alg \ref{alg:prediction} is used to generate a new datapoint.
The augmented data is appended to $D$ and the network is retrained.
This second part can be seen as a fine-tuning of the network parameter for data points that are close the prototype.
The data generation and retraining procedure is repeated until there is no change in network parameter $\parH$.
\begin{figure}[t]
\begin{algorithm}[H]
\begin{algorithmic}
	\State \textbf{Input}\\
	Model $\model(\parA,\parF)$\\
	Upper and lower parameter limits $L^{\t{up}} ,L^{\t{low}}$\\
	$\data=\emptyset$
	\vspace{.3\baselineskip}
	\State \textbf{// Generate an initial data set}
	\State $\t{\textbf{for} } d=1:N_{\t{data}}$
	\State $\quad (\parA, \parF) \sim \mathcal{U}(L^{\t{up}} ,L^{\t{low}})$
	\State $\quad (\depth,\pc) = \t{render}(\model(\parA, \parF))$
	\State $\quad \data = \data \cup \{(\depth, \pc, \parA, \parF)\}$
	\State $\t{\textbf{end}}$
	\vspace{.3\baselineskip}
	\State \textbf{// Train network}
	\State $\parH = \argmin{\parH} \sum_{i\in\data} ||  (\parA\iIdx, \parF\iIdx) - \nn(\depth\iIdx;\parH) ||^2$
	\vspace{0.6\baselineskip}	
	\State \textbf{// Generate data with model predictions}
	\State $N_{\t{pred}}=1$
	\Repeat
	\State $\t{\textbf{for} } d=1:N_{\t{aug}}$
	\State $\quad(\depth,\pc,\parA,\parF)=$ predict$(\depth\idx{d},\pc\idx{d},\parA\idx{d},\parH,N_{\t{pred}})$
	\State $\quad \data = \data \cup \{(\depth, \pc, \parA, \parF\idx{d})\}$
	\State $\t{\textbf{end}}$
	\State \textbf{// Retrain network}
	\State $\parH = \argmin{\parH} \sum_{i\in\data} ||  (\parA\iIdx, \parF\iIdx) - \nn(\depth\iIdx;\parH) ||^2$
	\vspace{.3\baselineskip}
	\State $N_{\t{pred}} = N_{\t{pred}} + 1$
	\Until no change in $\parH$\\
	\vspace{.3\baselineskip}
	\Return $\parH$
\end{algorithmic}
\caption{\!\!\textbf{:} Data Generation and Network Training}
\label{alg:training}
\end{algorithm}
\vspace{-2.5em}
\end{figure} 

\subsection{Network Architecture} 
\label{sub:network_architecture} 
\begin{figure*}
	\centering
	\small
\begin{tabular}{|l|c|c|c|c|c|c|c|c|c|c|}
	\hline
	\textbf{\specialcell{Scenario +\\Parameter}}
	& $L^\t{low}$
	& $L^\t{up}$
	& $N_\t{data}$
	& \textbf{\specialcell{Network\\architecture}}
	& \textbf{\specialcell{Baseline\\(Train)}}
	& \textbf{\specialcell{Baseline\\(Test)}}
	& \textbf{\specialcell{KMN \\(Train)}} 
	& \textbf{\specialcell{KMN \\(Test)}} 
	& \textbf{\specialcell{ICP \\(Train)}}
	& \textbf{\specialcell{ICP \\(Test)}}\\
	\hline	
box A & & & 40000 & [ 2  4  6  8 10] & 0.00452 & 0.00451 & \textbf{0.00143} & \textbf{0.00142} & 0.01668 & 0.01681\\
$\parA$: & & & & & & & & & & \\
x translation & -0.4 & 0.4 &  &  & 0.0021 & 0.0021 & 0.0007 & 0.0007 & 0.0074 & 0.0072 \\
y translation & -0.4 & 0.4 &  &  & 0.0024 & 0.0024 & 0.0007 & 0.0007 & 0.0093 & 0.0096 \\
\hline
box B & & & 60000 & [ 2  4  6  8 10] & 0.07125 & 0.06968 & \textbf{0.00734} & \textbf{0.00734} & 0.63701 & 0.62813\\
$\parA$: & & & & & & & & & & \\
x translation & -0.4 & 0.4 &  &  & 0.0095 & 0.0096 & 0.0016 & 0.0016 & 0.0102 & 0.0106 \\
y translation & -0.4 & 0.4 &  &  & 0.0116 & 0.0113 & 0.0018 & 0.0018 & 0.0128 & 0.0133 \\
z axis rotation & -1.0 & 1.0 &  &  & 0.0501 & 0.0488 & 0.0040 & 0.0040 & 0.6140 & 0.6042 \\
\hline
box C & & & 100000 & [ 4  8 10 12 14] & 0.18861 & 0.19106 & \textbf{0.04242} & \textbf{0.04295} & - & -\\
$\parA$: & & & & & & & & & & \\
x translation & -0.4 & 0.4 &  &  & 0.0135 & 0.0134 & 0.0029 & 0.0028 & - & - \\
y translation & -0.4 & 0.4 &  &  & 0.0131 & 0.0134 & 0.0028 & 0.0027 & - & - \\
z axis rotation & -1.0 & 1.0 &  &  & 0.0808 & 0.0843 & 0.0097 & 0.0109 & - & - \\
length scaling & -0.4 & 0.4 &  &  & 0.0421 & 0.0396 & 0.0138 & 0.0130 & - & - \\
height scaling & -0.5 & 1.5 &  &  & 0.0391 & 0.0403 & 0.0132 & 0.0135 & - & - \\
\hline
door & & & 100000 & [ 2  4  8 16 32] & 0.13852 & 0.13944 & \textbf{0.05253} & \textbf{0.05307} & - & -\\
$\parA$: & & & & & & & & & & \\
x translation & -0.8 & 0.8 &  &  & 0.0122 & 0.0124 & 0.0030 & 0.0029 & - & - \\
y translation & -0.8 & 0.8 &  &  & 0.0082 & 0.0085 & 0.0032 & 0.0032 & - & - \\
z axis rotation & -1.0 & 1.0 &  &  & 0.0198 & 0.0195 & 0.0053 & 0.0055 & - & - \\
\specialcellL{$\parF$:} & & & & & & & & & & \\
door height & -0.4 & 0.2 &  &  & 0.0150 & 0.0147 & 0.0120 & 0.0118 & - & - \\
door width & -0.2 & 0.2 &  &  & 0.0166 & 0.0169 & 0.0045 & 0.0045 & - & - \\
handle y & -0.0 & 0.0 &  &  & 0.0186 & 0.0190 & 0.0106 & 0.0107 & - & - \\
handle z & -0.1 & 0.1 &  &  & 0.0482 & 0.0485 & 0.0140 & 0.0145 & - & - \\
\hline
\end{tabular}
\caption{Results of experiment \ref{ssec:evaluation}.}
\label{fig:eval_results}
\vspace{-1.em}
\end{figure*} 
The function $f$ is parametrized as a multi-layer convolutional neural network.
The data consists of depth images with width $W=640$ and height $H=480$ and corresponding point clouds with $WH$ points.
The depth images are downsampled to a resolution of $128 \times 96$ before using them as an input to the network while the point cloud
dimensionality is kept the same.
This downsampling has two benefits: 1) A reduction of the amount of network parameter $\parH$; 2) The conversion from point clouds to depth images is better since there are fewer holes that might occur through scaling/rotation operations.
The depth values are normalized between a value of $0$ and $1$, where a depth value of $0$ belongs to the background and all other values to
the environment.
The basic structure of the network consists of $5$ convolution layers where each layer is followed by a max-pooling layer.
We use rectified linear unit activation function in the convolutional layers and a kernel size of $3\times 3$.
The number of channels of the convolutional layers is chosen dependent on the model complexity.
The last layer of the network is a linear layer that outputs the parameters $(\parA,\parF)$.
The loss function is the mean squared error that is optimized with the algorithm ADAM \cite{kingma2014adam}.

\section{Experiments} 
\label{sec:experiments} 
The proposed approach is evaluated in three experiments.
In the first experiment, the performance is compared to alternative strategies on different tasks with varying complexity.
The second experiment shows the adaption to real world sensor data and the third experiment demonstrates the transfer of a policy between different simulated door environments.
\subsection{Evaluation of Kinematic Morphing Networks}
\label{ssec:evaluation} 
\begin{figure}[t]
	\centering
\subfloat[][]{
	\includegraphics[width=.48\textwidth]{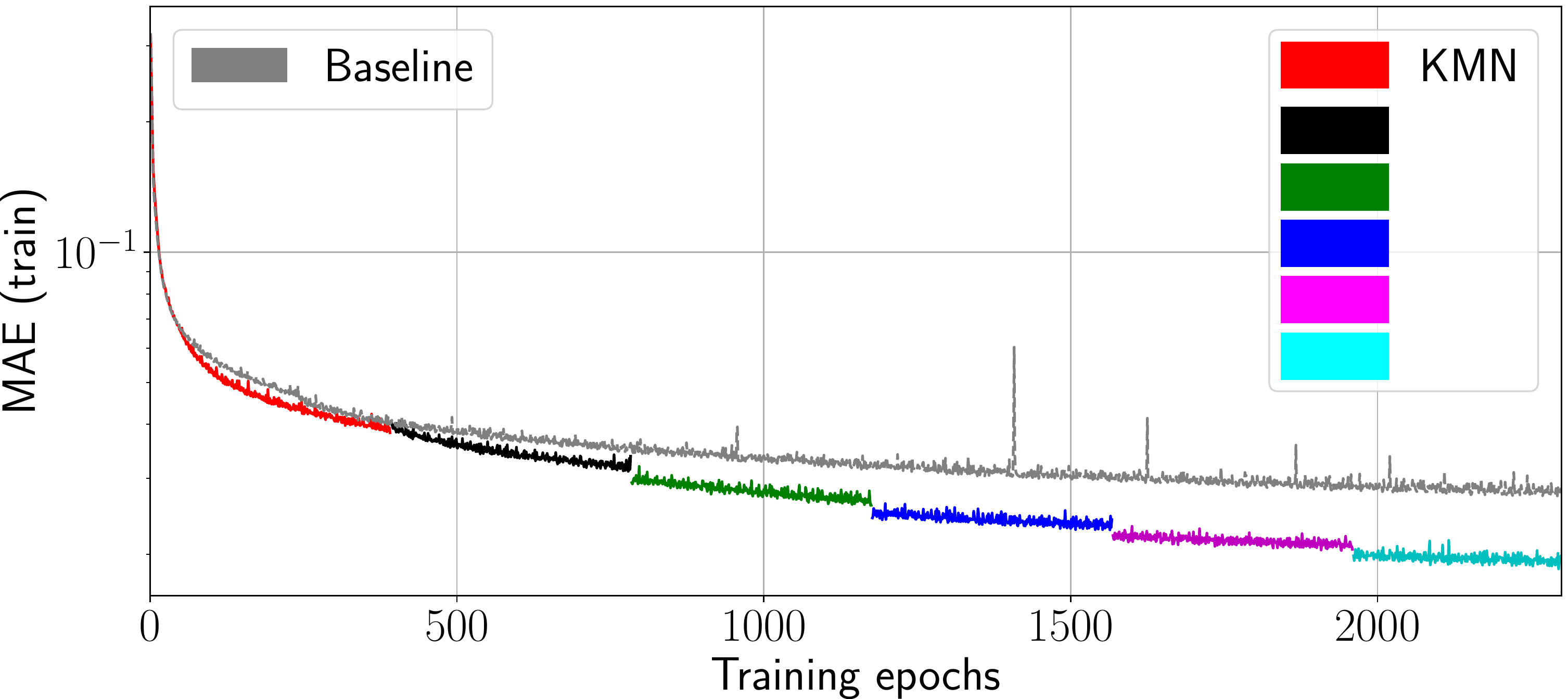}
		\label{fig:eval_train_error}
}

\vspace{-.5em}

\subfloat[][]{
	\includegraphics[width=.48\textwidth]{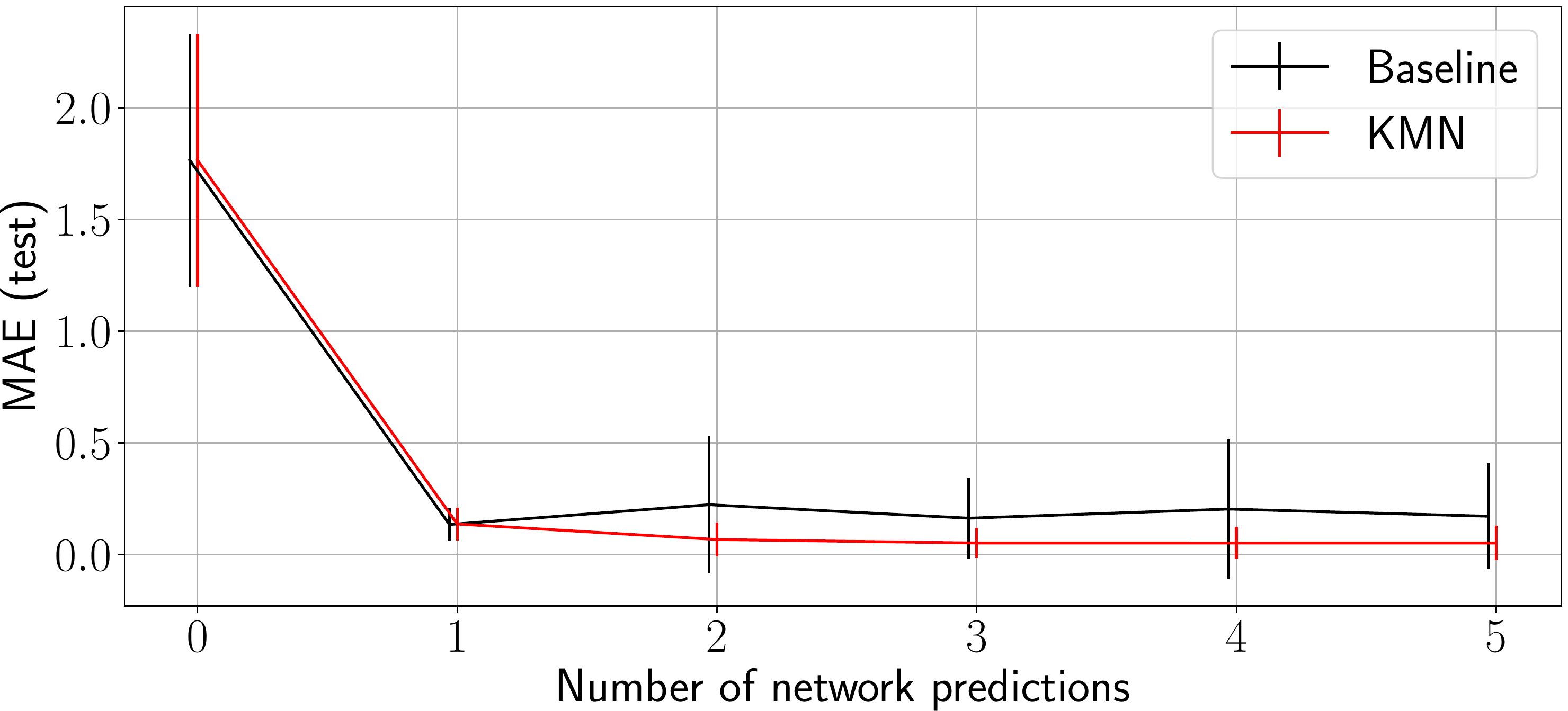}
	\label{fig:eval_mspe}
}
\vspace{-.5em}

\caption{The graphs in \protect\subref{fig:eval_train_error} show the training error of the Baseline and KMN variant. The alternating colors denote a new retraining iteration in \alg \ref{alg:training}. The plot in \protect\subref{fig:eval_mspe} shows the prediction error with standard deviation over multiple predictions with each network.}  
\end{figure}
\begin{figure*}[t]
	\centering
	\includegraphics[width=0.91\textwidth]{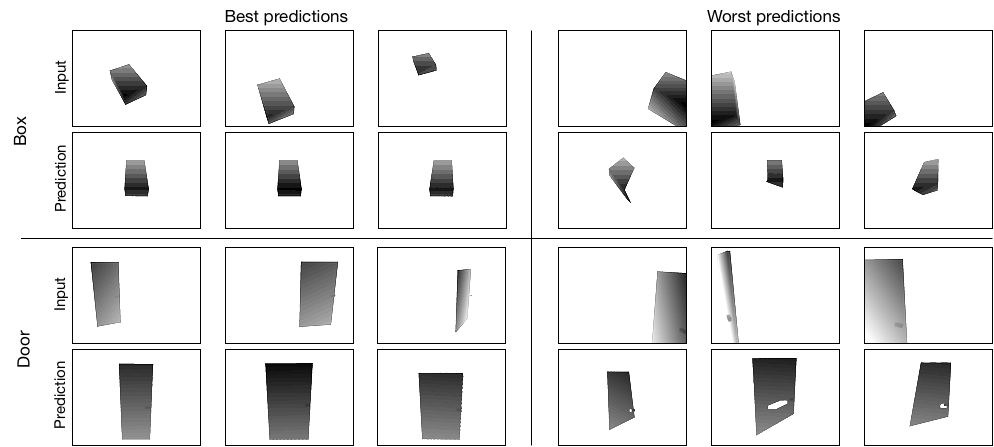}
	\caption{In the top row are the input depth images and on the bottom are the corresponding transformed images from the network predictions. The three samples on the left show the best predictions on the test dataset.}
	\label{fig:samples}
	\vspace{-.8em}
\end{figure*}

 
The prediction accuracy is evaluated on two tasks:
\begin{enumerate} 
	\item \textbf{box:} Three different box parametrizations are defined with varying complexity: A) \mbox{$n=2$}: the box is only translated along the
horizontal $x$ and $y$ direction; B) \mbox{$n=3$}: the box is additionally rotated around the $z$ axis; C) \mbox{$n=5$}: the box is additionally scaled in its
width and height. 
	\item \textbf{door:} This environment has \mbox{$n=3$} transformation parameters and \mbox{$m=4$} configuration parameters. The transformation parameters
$\parA$ are, similar to the box, the translation along $x$ and $y$ and the rotation around the $z$ axis. The configuration parameters $\parF$ are the size of
the door and the location of the door handle. The parametrization is sketched in \fig \ref{fig:door_setup}.
\end{enumerate}
\fig \ref{fig:samples} shows several depth images of both tasks.
The prototype is, in both tasks, defined at $\parA_0=\v 0$ and $\parF_0=\v 0$, which corresponds to the configuration where the object is directly in front
of the robot.
We compare different algorithms on both environments that predict the morphing parameters $(\parA, \parF)$ from depth images and point clouds. The algorithms
are:
\begin{itemize} 
	\item \textbf{Kinematic Morphing Network (KMN):} This is the method proposed in this paper (see \sect \ref{sec:methodology}) with $N_\t{pred}=5$ number of predictions.
	\item \textbf{Baseline:} Uses the same neural network as KMN. However, the network is trained only on the initially generated data without retraining and applied with a single prediction step ($N_\t{pred}=1$).
	\item \textbf{Iterative Closest Point (ICP):} The iterative closest point algorithm \cite{chen1992object} with $100$ iterations.
\end{itemize}
The results of this experiment are shown in the table in \fig \ref{fig:eval_results}.
The table lists the environment parameters, model architectures, and prediction error for all tasks and algorithms.
The network architecture describes the number of channels in the $5$ convolutional layers.
The data is split into $80\%$ train and $20\%$ test data.
The initially generated amount of data $N_\t{data}$ and the network architecture is chosen heuristically according to the complexity of the task.
The number of augmented data points $N_\t{aug}$ is set to $20\%$ of $N_\t{data}$.
The reported error metric is the mean absolute error over $1000$ data points and reported on the train and test set for each algorithm. 

The results indicate that the proposed approach KMN achieves the lowest prediction errors on all tasks.
The difference between KMN and Baseline comes through the iterative prediction and retraining mechanisms, since both variants use the same network architecture.
\fig \ref{fig:eval_train_error} shows the training error of the Baseline and KMN variants on the box environment C.
A different color denotes a new iteration of the KMN retraining loop in \alg \ref{alg:training}.
The data augmentation of KMN leads to a faster decrease of the training error.
\fig \ref{fig:eval_mspe} shows the prediction error with standard deviation of Baseline and KMN over number of predictions on the test set.
The first prediction of both networks achieves a similar error.
However, the KMN method improves the prediction by applying the network multiple times.
After $3$ iterations there is no significant change in the accuracy anymore.
The prediction error of the baseline increases since it only was trained on the initial dataset.
This shows that the retraining mechanism is necessary in order to apply the network iteratively.
The ICP algorithm was applied on box A and box B since they do not have any configuration parameter.
ICP achieves a reasonable performance on the translation parameter of both environments.
However, ICP had difficulties on the rotation parameter since it sometimes did not detect the correct rotation direction or led to rotations
that flipped the box.

\fig \ref{fig:samples} shows different samples of the dataset (top row) with the corresponding network prediction (bottom row) separated in best and worst predictions.
The worst predictions occurred when the box/door was only partially observed.
This makes sense since it is difficult to estimate the height of a door when it is not fully visible.
The samples also show that the transformation of point clouds and the subsequent rendering can lead to holes in the depth image.
However, since the KMN network also has such data points in the training phase, it could handle them better than the Baseline.
The morphing transformations of the door and box task are shown in the appended video.

\begin{figure}[t] 
	\centering
	\includegraphics[width=.45\textwidth]{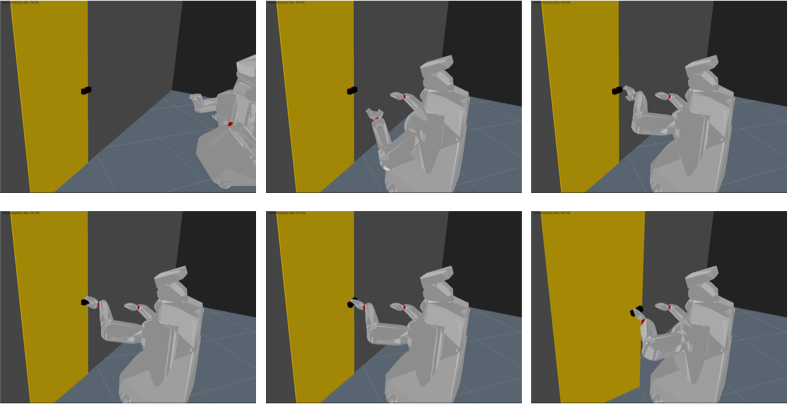}
	\caption{Sequence of door a opening motion.}
	\label{fig:door_skill}
	\vspace{-.5em}
\end{figure}

\subsection{Evaluation on Real Sensor Data} 
\label{sub:evaluation_on_real_sensor_data} 
\begin{figure*}[t]
	\centering
	\includegraphics[width=.85\textwidth]{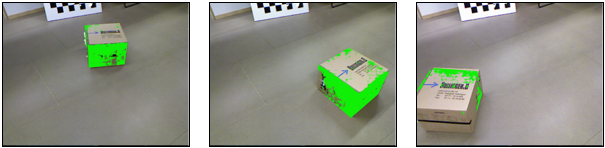}
	\caption{Kinematic morphing network predictions (green box) overlaid with point clouds from a Kinect camera.}
	\label{fig:kinect_predictions}
	\vspace{-1.5em}
\end{figure*}
In this experiment, we evaluate how the KMN model performs on real sensor data.
The trained model of the box C task is used on data recorded with a Kinect v1 camera.
The point clouds are recorded with the IR depth-finding camera of the Kinect.
We tried to reproduce the simulated environment in the real world (e.g., same camera pose).
The point clouds are preprocessed by transforming the points from camera into world frame and removing the points that do not belong to the object.
We put the box at $15$ different locations inside the field of view of the camera.
The network was able to predict morphing parameters for all $15$ samples that transform the observed box close to the prototype.
The achieved accuracy was lower and the number of network predictions $N_\t{pred}$ until convergence was slightly higher in comparison to simulation data. 
\fig \ref{fig:kinect_predictions} shows different point clouds overlaid with their predicted box location (green box).

\subsection{Skill Transfer on the Door Task} 
\label{ssec:door_transfer} 
We use the trained KMN model to transfer a skill policy between different doors.
The policy is defined on the kinematic model $\model(\parA, \parF)$ as a constrained optimization problem (see \sect \ref{sub:policy_representation}).
The robot is a PR2 and the trajectory $\trajI$ consists of $T=200$ configurations of the robot base (3 dof), left arm (7 dof), gripper (1 dof), and door (2 dof).
The features $\v\Phi$ of the cost function are defined as follows:
\begin{itemize}
	\item Base pose in front of the door.
	\item Pre-grasp pose of gripper in front of door handle.
	\item Target state of handle joint.
	\item Target state of door joint.
\end{itemize}
The equality constraint $\v h$ consists of a feature that describes the contact between door handle and gripper.
Specifically, two points are defined on the door handle and on the robot gripper.
The constraint measures the difference between a point pair that should be zero during the manipulation.
Further constraints are defined to avoid collisions and to fix joints when they are not being manipulated.
Figure \ref{fig:door_skill} shows the skill on an instance of the door environment.
The skill is also shown in the video in the supplementary material.

\section{Conclusion} 
\label{sec:conclusion}
We introduced kinematic morphing networks to transfer manipulation skills between different environments.
Kinematic morphing networks extract parameters from depth images and are trained on data generated with a simulator.
The conversion between point clouds and depth images allows to apply the network in an iterative manner, which increases the overall accuracy.
We demonstrated the network performance on real sensor data and the transfer of a skill with a motion planning method.
 
\bibliographystyle{IEEEtran}
\bibliography{refs}

\end{document}